\documentclass[conference]{IEEEtran}
\IEEEoverridecommandlockouts
\usepackage{booktabs} 
\usepackage{float}
\usepackage{subcaption}
\usepackage{cite}
\usepackage{amsmath,amssymb,amsfonts}
\usepackage{algorithmic}
\usepackage{graphicx}
\usepackage{textcomp}
\usepackage{xcolor}
\def\BibTeX{{\rm B\kern-.05em{\sc i\kern-.025em b}\kern-.08em
    T\kern-.1667em\lower.7ex\hbox{E}\kern-.125emX}}
\begin{document}

\title{Graph Representation Learning Strategies for Omics Data:\\ A Case Study on Parkinson's Disease }

\author{\IEEEauthorblockN{Elisa Gómez de Lope\textsuperscript{*}\thanks{*Corresponding author. Email: elisa.gomezdelope@uni.lu}}
\IEEEauthorblockA{\textit{Department of Engineering} \\
\textit{University of Luxembourg}\\
elisa.gomezdelope@uni.lu
}
\and
\IEEEauthorblockN{Saurabh Deshpande}
\IEEEauthorblockA{\textit{Department of Engineering} \\
\textit{University of Luxembourg}\\
saurabh.deshpande@uni.lu}
\and
\IEEEauthorblockN{Ramón Viñas Torné}
\IEEEauthorblockA{\textit{Dept. of Computer and Communication Sciences} \\
\textit{École polytechnique fédérale de Lausanne (EPFL)}\\
ramon.vinastorne@epfl.ch}
\and
\IEEEauthorblockN{Pietro Liò}
\IEEEauthorblockA{\textit{Department of Computer Science} \\
\textit{University of Cambridge}\\
pl219@cam.ac.uk}
\and
\IEEEauthorblockN{Enrico Glaab\textsuperscript{1}\thanks{1 On behalf of the NCER-PD Consortium}}
\IEEEauthorblockA{\textit{Luxembourg Center for Systems Biomedicine} \\
\textit{University of Luxembourg}\\
enrico.glaab@uni.lu}
\and
\IEEEauthorblockN{Stéphane P. A. Bordas}
\IEEEauthorblockA{\textit{Department of Engineering} \\
\textit{University of Luxembourg}\\
stephane.bordas@uni.lu}

}

\maketitle
 
\begin{abstract}
Omics data analysis is crucial for studying complex diseases, but its high dimensionality and heterogeneity challenge classical statistical and machine learning methods. Graph neural networks have emerged as promising alternatives, yet the optimal strategies for their design and optimization in real-world biomedical challenges remain unclear. This study evaluates various graph representation learning models for case-control classification using high-throughput biological data from Parkinson's disease and control samples. We compare topologies derived from sample similarity networks and molecular interaction networks, including protein-protein and metabolite-metabolite interactions (PPI, MMI). Graph Convolutional Network (GCNs), Chebyshev spectral graph convolution (ChebyNet), and Graph Attention Network (GAT), are evaluated alongside advanced architectures like graph transformers, the graph U-net, and simpler models like multilayer perceptron (MLP).

These models are systematically applied to transcriptomics and metabolomics data independently. Our comparative analysis highlights the benefits and limitations of various architectures in extracting patterns from omics data, paving the way for more accurate and interpretable models in biomedical research.

\end{abstract}

\begin{IEEEkeywords}
Graph neural networks, transformers, machine learning, omics, transcriptomics, metabolomics, Parkinson's disease
\end{IEEEkeywords}
\vspace{-0.5cm}
\section{Introduction}
Analyzing omics data in complex diseases like Parkinson's disease (PD) is challenging due to high dimensionality, noise, heterogeneity, and typically small sample sizes. Recently, graph representation learning methods combining machine learning and graph theory, have reported potential in tackling these challenges by exploring relational information and capturing structural and functional interactions \cite{b7,b8,b9,b10}. Graph Neural Networks (GNNs) propagate feature information through graph structures, making them particularly well-suited for biomedical data where complex interrelationships can be intuitively represented as networks. 

Previous work, such as that of Chereda et al. \cite{b1, b2}, has used GNNs to predict disease outcomes using a PPI network. Other approaches have mapped multi-omics data into patient similarity networks \cite{b3, b4, b5}, but acquiring such data from the same cohort is often impractical for PD \cite{b6}. Our study aims to address this gap by comparing different graph representation learning methods to identify patterns within individual omics data in PD. The findings may guide future research in modeling omics data and uncover fingerprints relevant to PD, contributing to a deeper understanding of its pathology.

\section{Materials and Methods}
\vspace{-0.2cm}
\subsection{Datasets}
\vspace{-0.1cm}
Both sample similarity and molecular interaction networks modelling pipelines were trained and evaluated in two independent cohorts: PPMI (whole blood transcriptomics) \cite{b11} and LUXPARK (blood plasma metabolomics) \cite{b12}. After the pre-processing steps described in Appendix \ref{data_preprocessing}, the PPMI dataset contained 14,548 gene expression features for 378 samples (189 controls and 189 PD patients), while the LUXPARK set contained 1,079 metabolite profiles for 1,136 subjects (590 controls and 546 PD patients). Pre-processing of metabolomics data only partially mitigated the medication effect, potentially obscuring the biological interpretation of the results from LUXPARK data. 
\vspace{-0.1cm}
\subsection{Modelling Sample Similarity Networks}
\vspace{-0.1cm}
Sample-sample similarity networks are an emerging paradigm in modelling omics data \cite{b3, b4, b5}. Here, an end-to-end modeling pipeline was developed, trained, and evaluated in a 10-fold cross-validation for PD case-control node classification on LUXPARK (metabolomics) and PPMI (transcriptomics) data. The pipeline includes 5 steps: data scaling, feature selection with LASSO \cite{{b14}}, building a sample-sample similarity graph, the formulation of the GNN model for the binary node classification task, and a post-hoc explainer module using GNN-Explainer \cite{b15} (see schema in Figure \ref{fig:ssn_schema} of Appendix \ref{schemas}).

The topology of the network is defined with an adjacency matrix $A$, which is constructed by finding pairwise cosine similarity scores among samples that are larger than a threshold hyperparameter $s$ (eq. \ref{eqn:graph_adj}) \cite{b16}:


\vspace{-0.6cm}
\begin{equation}
\label{eqn:graph_cosine_similarity}
\text{cosine similarity} (i, j) = cs({{x}}_{i},{{x}}_{j}) = \cos (\theta ) = \dfrac {x_i \cdot x_j} {\left\| x_i\right\| _{2}\left\| x_j\right\| _{2}} 
\end{equation}
\vspace{-0.15cm}
where $\left\| x_i \right\|_{2}$ is the 2-norm of the feature vector $x_i$ of sample $i$.
\vspace{-0.05cm}
\begin{equation}
\label{eqn:graph_adj}
    {A}_{ij}=\left\{\begin{array}{ll}cs({{x}}_{i},{{x}}_{j}),&{\rm{if}}\ i\, \ne\, j\ {\rm{and}}\ cs({{x}}_{i},{{x}}_{j})\ge s \\ 0,&\,{\text{otherwise}}\hfill\,\end{array}\right.
\end{equation}

\subsection{Modelling Molecular Interactions Networks}
Molecular interaction networks represent relationships between molecular entities. A modeling pipeline based on graph representation learning was implemented to classify PD cases and controls mapping a PPI on the transcriptomics data from PPMI, and a MMI on the metabolomics data from LUXPARK independently. The pipeline includes several steps: constructing the PPI from STRING (v12.0) \cite{b17}, or MMI from STITCH (v5.0) \cite{b14}, adjusting the feature matrix of omics profiles $X$ to include only molecules represented in the network, scaling the data, formulating the GNN model as a binary graph classification task, and analyzing explainability with GNN-Explainer (a schema is available in Figure \ref{fig:min_schema} of Appendix \ref{schemas}).

Notably, not all transcriptomics and metabolomics features mapped to the PPI and MMI, respectively. In PPMI, the final graph $G=(E,V)$ included 5,848 nodes, and the feature matrix $X$ was adjusted accordingly. The MMI network for LUXPARK included only 474 vertices (44\% of the initial feature set), despite efforts in matching features via multiple chemical identifiers including PubChem, KEGG, ChEBI, CAS, SMILES and inchikeys.
\vspace{-0.4cm}
\subsection{Comparative analysis and evaluation}
\vspace{-0.1cm}

We evaluated the performance of several models, including GCN, ChebyNet, and GAT, applied to molecular interaction networks (graph classification) and sample similarity networks (node classification) for PD vs. control classification on PPMI and LUXPARK datasets. Their performance was also compared against that of an MLP \cite{b18}, considering factors like feature selection and model depth. Additionally, we assessed the performance of various state-of-the-art GNN and transformer-based approaches, including Graph U-Net \cite{b19} and graph transformers (GPST \cite{b20}, GTC \cite{b21}). Finally, the models were compared with similarity-based versus random edges.

Unless specifically mentioned, all models use 2 layers of their respective transformation followed by ReLU activation \cite{b22}.

\section{Results and discussion}
\vspace{-0.1cm}
\subsection{Sample similarity versus molecular interaction networks}
\vspace{-0.1cm}

We compared the performance of GCN, ChebyNet, and GAT models in the two modeling pipelines: using sample similarity networks (SSNs) and molecular interaction networks (MINs). We found that SSNs consistently outperformed MINs, particularly in the LUXPARK dataset. We hypothesize that this discrepancy is attributed to the incompleteness of MINs, especially that of metabolite-metabolite interactions, and the potential irrelevance of healthy individual-based MINs to PD pathology.

\vspace{-0.1cm}
\subsection{The power of the graph structure: MLP vs GNNs} 
\vspace{-0.1cm}
GNNs generally outperformed the MLP, indicating that GNNs can more effectively integrate information, capturing complex dependencies and interactions in the omics profiles that the MLP might miss.

\vspace{-0.1cm}
\subsection{The effect of feature selection and depth of the network}
\vspace{-0.1cm}
The ablation experiments show that the incorporation of LASSO-based feature selection consistently improved model performance, addressing the challenge of modeling high-dimensional high-throughput biological data with limited sample sizes.

Increasing the number of layers did not improve model performance, likely due to overfitting given the limited sample size.

\vspace{-0.05cm}
\subsection{Transformers generally outperform traditional GNNs}
\vspace{-0.1cm}

The graph transformer architectures studied here showed higher average cross-validated performance than "traditional" GNN layers (GCN, GAT, Chebyshev) in both datasets. Among traditional layers, \textit{ChebyNet} showed slightly higher performance. Other architectures like Graph U-Net generally underperformed compared to traditional GNNs. Graph transformers seem to be more expressive, leveraging the sample-similarity through the edge attributes. 
\vspace{-0.15cm}
\subsection{Inductive biases: similarity-based versus random edges}
\vspace{-0.1cm}

In PPMI, models with similarity-based edges reported higher performance metrics than those with random edges. Graph transformers were less sensitive to edge type, likely due to their ability to infer these nuances through the attention mechanism. In LUXPARK, models performed comparably regardless of edge type, suggesting that larger sample size may reduce the need for inductive biases. Additionally, sparser networks like SSNs may limit the flow of information and hinder learning rich embeddings.

\vspace{-0.1cm}
\begin{table}[htbp]
  \caption{Cross-validated performance for molecular interaction networks in PPMI and LUXPARK.}
    \begin{tabular}{lcccc}
          & \multicolumn{2}{c}{\textbf{PPI network (PPMI)}} & \multicolumn{2}{c}{\textbf{MMI network (LUXPARK)}} \\
          & \textbf{AUC} & \textbf{F1} & \textbf{AUC} & \textbf{F1} \\
    \midrule
    ChebyNet & 0.47 ± 0.08 & 0.38 ± 0.29 & 0.54 ± 0.07	& 0.36 ± 0.26\\
    GCN   & 0.52 ± 0.05 & 0.48 ± 0.28 & 0.48 ± 0.07	& 0.39 ± 0.14\\
    GAT   & 0.54 ± 0.04 & 0.52 ± 0.22 & 0.55 ± 0.06 & 0.52 ± 0.05\\
    \bottomrule
    \end{tabular}%
  \label{tab:GRAPH_performance_TRANSC}%
\vspace{-1.2em}
\end{table}%

\vspace{-0.2cm}
\begin{table}[htbp]
  \centering
  \caption{Cross-validated performance for sample similarity networks in PPMI and LUXPARK.}
    \begin{tabular}{p{1.9cm}cccc}
          & \multicolumn{2}{c}{\textbf{PPMI}} & \multicolumn{2}{c}{\textbf{LUXPARK}} \\
     & \textbf{AUC}   & \textbf{F1}    & \textbf{AUC}   & \textbf{F1} \\
    \midrule
    ChebyNet & \textbf{0.58 ± 0.08} & 0.52 ± 0.18 &\textbf{ 0.83 ± 0.05} & \textbf{0.73 ± 0.06} \\
    GAT   & 0.56 ± 0.07 & \textbf{0.55 ± 0.06} & 0.79 ± 0.04 & 0.71 ± 0.05 \\
    GCN   & 0.56 ± 0.11 & 0.54 ± 0.09 & 0.81 ± 0.04 & 0.72 ± 0.05 \\
    \midrule
    MLP   & 0.55 ± 0.09 & 0.55 ± 0.07 & 0.78 ± 0.05 & 0.68 ± 0.06 \\
    \midrule
    Graph U-Net & 0.54 ± 0.11 & 0.47 ± 0.2 & 0.74 ± 0.19 & 0.69 ± 0.12 \\
    \midrule
    GPST\textsubscript{GINE} & \textbf{0.6 ± 0.07} & 0.54 ± 0.07 & 0\textbf{.84 ± 0.05} & \textbf{0.76 ± 0.06} \\
    GTC & 0.59 ± 0.09	& \textbf{0.55 ± 0.11} &	0.82 ± 0.06	 & 0.73 ± 0.07 \\
    \midrule
    ChebyNet\textsubscript{No LASSO} & 0.54 ± 0.08 & 0.52 ± 0.1 & 0.82 ± 0.06 & 0.75 ± 0.06 \\
    GAT\textsubscript{No LASSO} & 0.54 ± 0.11 & 0.53 ± 0.14 & 0.69 ± 0.05 & 0.65 ± 0.07 \\
    GCN\textsubscript{No LASSO} & 0.5 ± 0.09 & 0.43 ± 0.16 & 0.76 ± 0.03 & 0.7 ± 0.04 \\
    \midrule
    ChebyNet\textsubscript{10 Layers} & 0.54 ± 0.08 & 0.52 ± 0.08 & 0.82 ± 0.05 & 0.74 ± 0.05 \\
    ChebyNet\textsubscript{50 Layers} & 0.51 ± 0.09 & 0.49 ± 0.15 & 0.69 ± 0.08 & 0.61 ± 0.07 \\
    GAT\textsubscript{10Layers} & 0.49 ± 0.09 & 0.48 ± 0.08 & 0.67 ± 0.13 & 0.63 ± 0.14 \\
    GAT\textsubscript{50 Layers} & 0.53 ± 0.09 & 0.51 ± 0.1 & 0.59 ± 0.07 & 0.51 ± 0.16 \\
    GCN\textsubscript{10 Layers} & 0.55 ± 0.08 & 0.51 ± 0.09 & 0.69 ± 0.15 & 0.61 ± 0.12 \\
    \midrule
    ChebyNet\textsubscript{random E} & 0.51 ± 0.15 & 0.46 ± 0.18 & \textbf{0.84 ± 0.05} & 0.74 ± 0.04 \\
    GAT\textsubscript{random E} & 0.53 ± 0.07 & 0.53 ± 0.08 & 0.8 ± 0.07 & 0.7 ± 0.05 \\
    GCN\textsubscript{ramdom E} & 0.53 ± 0.09 & 0.49 ± 0.1 & 0.64 ± 0.13 & 0.6 ± 0.08 \\
    GPST\textsubscript{random E} & \textbf{0.59 ± 0.08} & \textbf{0.61 ± 0.05} & \textbf{0.84 ± 0.07} & \textbf{0.75 ± 0.06} \\
    GTC\textsubscript{random E} & 0.56 ± 0.09 & 0.55 ± 0.1 & \textbf{0.84 ± 0.03} & 0.73 ± 0.05 \\
    \bottomrule
    \end{tabular}%
  \label{tab:cv_ssn}%
\end{table}
\vspace{-0.5cm}
\subsection{Biological Insights}
\vspace{-0.2cm}
Both SSN and MIN modeling pipelines identified genes and metabolites involved in the mitochondrial shuttle of acylcarnitines for fatty acid $\beta$-oxidation as relevant, including genes \textit{SLC25A20} and \textit{CPT1A}, and glutarylcarnitine (C5-DC). These preliminary findings align with previous work \cite{b23}\cite{b24} on associations between mitochondrial fatty acids and PD.

\vspace{-0.2cm}
\section{Future work}
\vspace{-0.2cm}
We plan to further extract biological interpretations in the context of PD by identifying relevant molecular subnetworks and performing functional enrichment analysis. In addition, we will assess the statistical significance of our comparative analysis.

\vspace{-0.2cm}
\section*{Acknowledgements}
\vspace{-0.18cm}
We acknowledge support by Luxembourg's National Research Fund (FNR) through the grants no. FNR/NCER13/BM/11264123 and C20/MS/14782078/QuaC, and the Marie Sklodowska-Curie grant
agreement No. 764644.

%
%



\appendix
\vspace{-0.2cm}

\subsection{Data pre-processing}
\label{data_preprocessing}

The transcriptomics data from PPMI was quantified by the STAR + featureCounts method, and, a low expression filter was applied, resulting in 23,253 features. Metabolomics data from the LuxPARK cohort were measured and initially pre-processed by Metabolon®, and the log-transformed batch normalized peak-area data was used. Since levodopa (L-DOPA) dopaminergic medication has a pronounced effect on the metabolism of blood, we additionally filtered out all metabolites from the data with a minimum absolute Pearson correlation of 0.2 with the L-DOPA metabolite 3-O-methyldopa (also known as 3-OMD or 3-methoxytyrosine) and all metabolites belonging to tyrosine metabolism and tryptophan metabolism prior to the analysis, resulting in 86 (5.77\%) metabolites being eventually removed from the feature space.

Both datasets underwent unsupervised filtering (low-variance and highly correlated features were removed) prior to the sample-similarity network modelling pipeline, while transcriptomics data from the PPMI also underwent these unsupervised filters prior to the molecular interaction networks modelling to reduce the otherwise large feature space and mitigate a potential over-squashing effect after recursive aggregation of node features, creating a more compact and informative PPI network.

\subsection{Schemas of SSN and MIN modeling pipelines}
\label{schemas}
\vspace{-0.4cm}
\begin{figure}[H]
    \centering
    \includegraphics[clip, trim=1cm 0.5cm 1cm 0.5cm, width=1.025\linewidth]{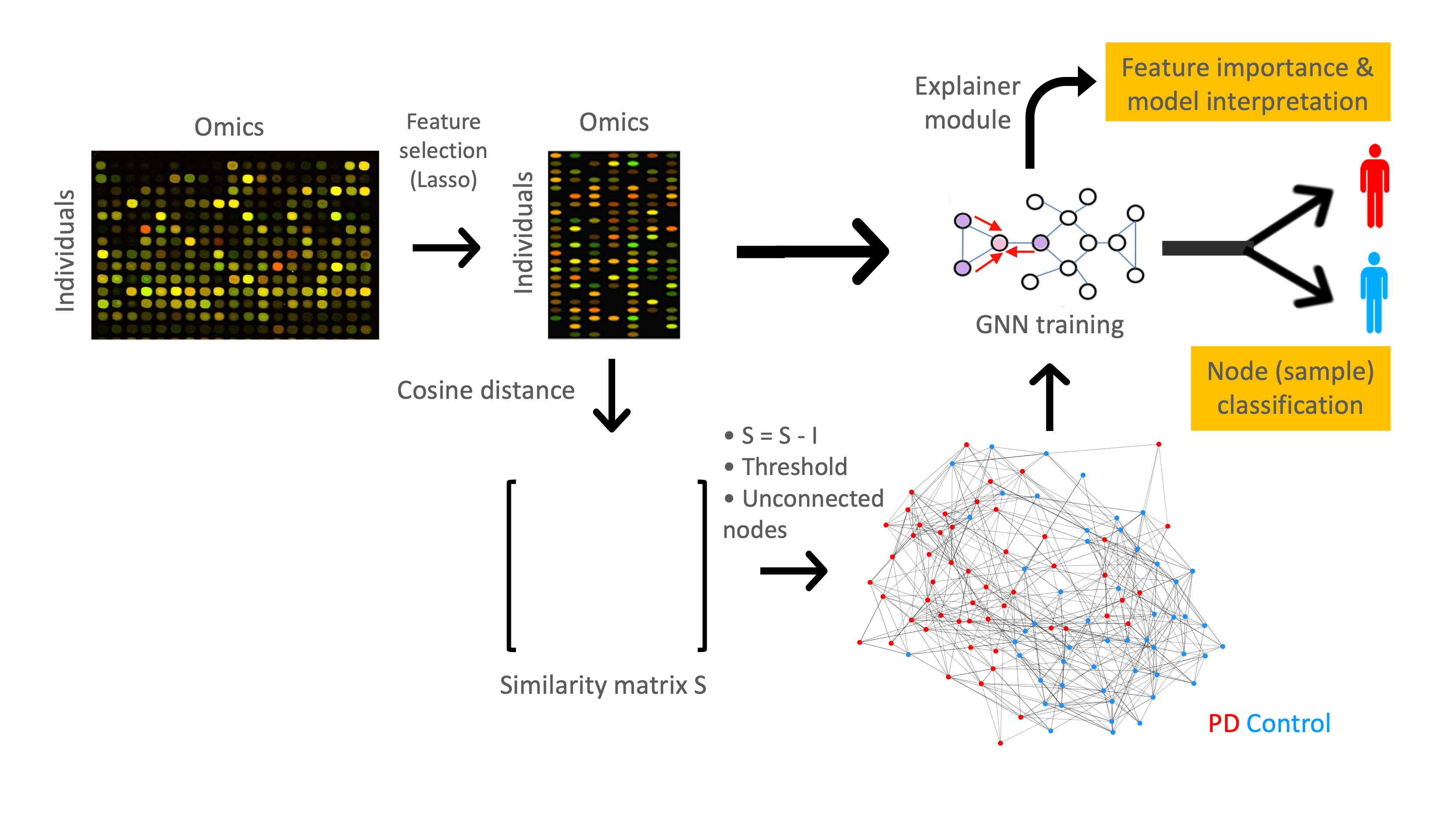}
   \captionsetup{font=footnotesize, skip=-0.1cm}
    \caption{Schema of graph representation learning using sample-sample similarity networks modeling pipeline.}
    \label{fig:ssn_schema}
\end{figure}
\vspace{-0.5cm}
\begin{figure}[H]
    \centering
    \includegraphics[clip, trim=0.1cm 0.1cm 0.1cm 0.17cm,width=1.025, width=1\linewidth]{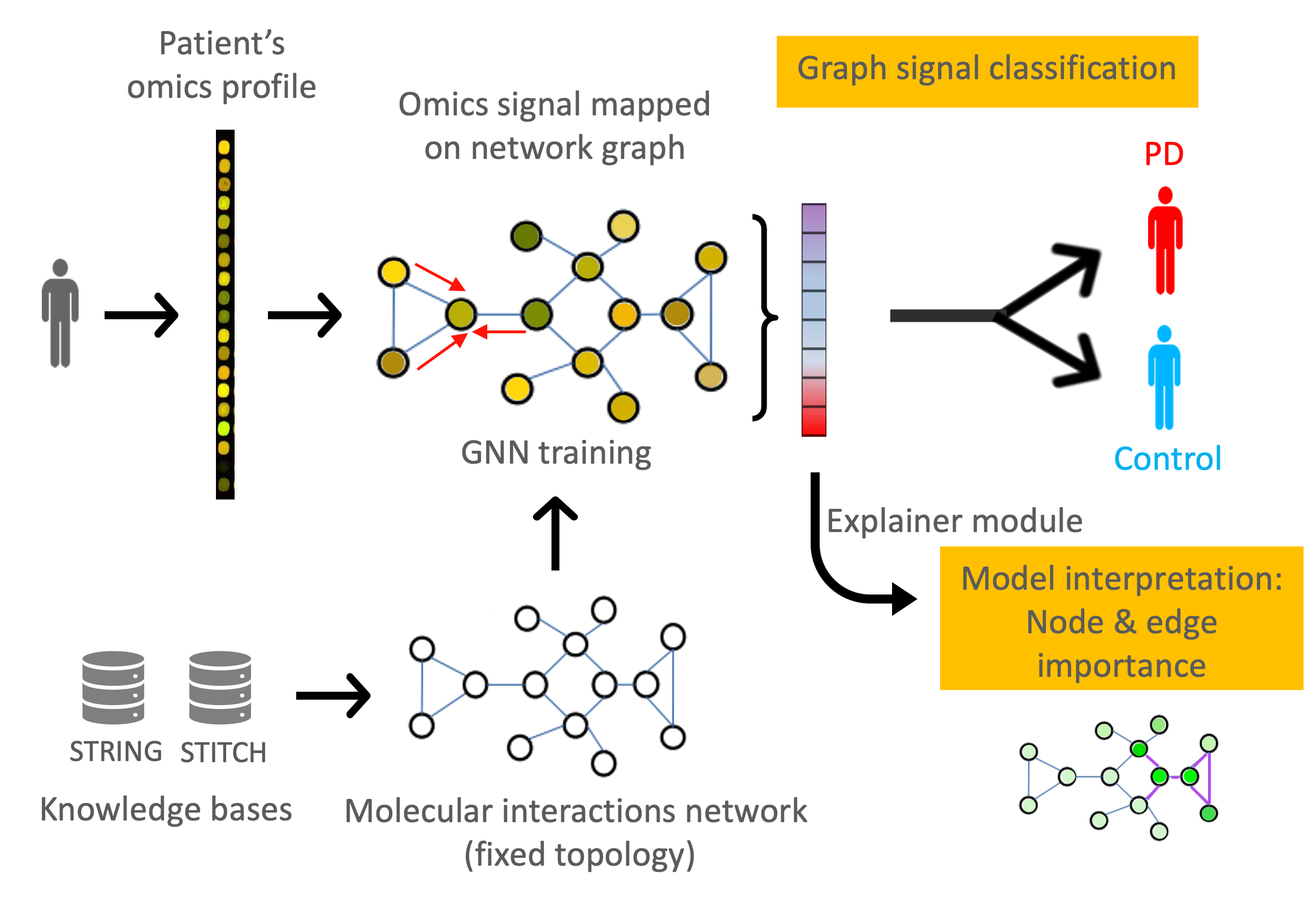}
   \captionsetup{font=footnotesize}
    \caption{Schema of graph representation learning using molecular interaction networks modeling pipeline.}
    \label{fig:min_schema}
\end{figure}

\end{document}